\documentclass[10pt,twocolumn,letterpaper]{article}

\usepackage[pagenumbers]{iccv} 
\usepackage{placeins}

\definecolor{iccvblue}{rgb}{0.21,0.49,0.74}
\usepackage[pagebackref,breaklinks,colorlinks,allcolors=iccvblue]{hyperref}

\def\paperID{*****} 
\def\confName{ICCV}
\def\confYear{2025}

\title{Simulating Refractive Distortions and Weather-Induced Artifacts for Resource-Constrained Autonomous Perception}

\author{
Moseli Mots’oehli$^{1,2, }$\thanks{Corresponding authors: \textit{moselim@hawaii.edu, huaijin@hawaii.edu}}, 
Feimei Chen$^1$, Hok Wai Chan$^1$, Itumeleng Tlali$^2$, Thulani Babeli$^2$,\\
Kyungim Baek$^1$, Huaijin Chen$^{1,*}$
\\
$^1$University of Hawai‘i at Mānoa \quad $^2$MindForge AI, South Africa
}

\begin{document}
\maketitle

\captionsetup{skip=1pt}
\begin{figure*}[!thb]
  \centering
  \includegraphics[width=\textwidth,trim = 0 1.3cm 0 0,   
    clip]{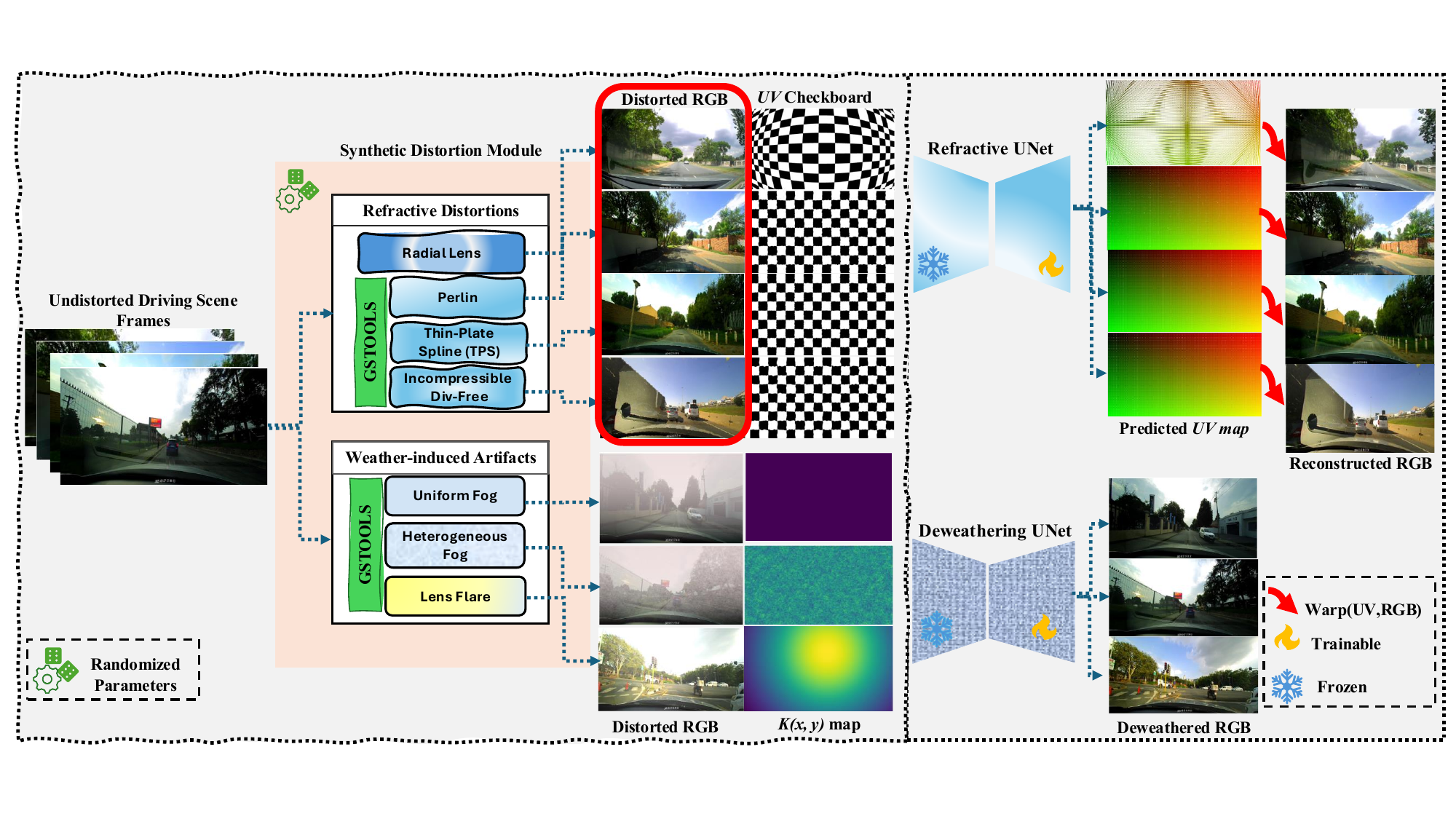}
  \caption{Overview of our two‐stage restoration pipeline. The Synthetic Distortion Module applies random refractive distortions and weather-induced artifacts described in Section \ref{sec:method}. The Refractive U-Net predicts a dense UV displacement field that's used to correct the distortion in the input image. The Deweathering U-Net removes weather-induced artifacts from the input image. The two models are trained separately.}
  \label{fig:framework}
\end{figure*}

\begin{abstract}
The scarcity of autonomous vehicle datasets from developing regions, particularly across Africa’s diverse urban, rural, and unpaved roads, remains a key obstacle to robust perception in low-resource settings. We present a procedural augmentation pipeline that enhances low-cost monocular dashcam footage with realistic refractive distortions and weather-induced artifacts tailored to challenging African driving scenarios. Our refractive module simulates optical effects from low-quality lenses and air turbulence, including lens distortion, Perlin noise, Thin-Plate Spline (TPS), and divergence-free (incompressible) warps. The weather module adds homogeneous fog, heterogeneous fog, and lens flare. To establish a benchmark, we provide baseline performance using three image restoration models. To support perception research in underrepresented African contexts, without costly data collection, labeling, or simulation, we release our distortion toolkit, augmented dataset splits, and benchmark results.
\end{abstract}    
\section{Introduction}
\label{sec:intro}
Autonomous driving research has advanced rapidly in urban and highway environments in developed countries, yet datasets from developing countries, particularly those in Africa, remain scarce \cite{motsoehli2022public}. Deploying delicate multi-modal sensor platforms and launching large-scale data collection and labeling efforts or performing high-budget simulations in these regions pose significant challenges. On the contrary, low-cost dashcams casually recording daily driving are becoming increasingly common. 
Even where such footage exists, it remains insufficient to capture the full diversity of African driving scenarios, including optical distortions from a wide range of camera models, turbulence-induced refraction, and complex weather variability. This paper addresses these gaps by procedurally augmenting real-world African dashcam footage with refractive distortions and weather-induced artifacts. With limited resources, we aim to generate sufficient training data as a first step toward enabling practical autonomous driving solutions in the Global South. Specifically, our contributions are as follows:
\begin{itemize}

  \item \textbf{Scalable refractive distortion simulation in low-resource settings:} Our synthetic distortion and artifact pipeline avoids costly hardware and fieldwork as well as high-budget physics-based rendering, enabling affordable dataset expansion in low-resource settings that models the most common refraction distortions.

  \item \textbf{Context-aware weather artifact simulation:} We simulate harsh driving conditions, such as Cape Town fog or Limpopo sun glare, through a modular, efficient pipeline that produces paired clean/corrupted frames.

  \item \textbf{Un-distortion and de-weathering baselines and benchmarks:} We show recovery of clean images from above mentioned artifacts and distortion using three baseline models: a ResUnet (CNN encoder), SegFormer-B1 (Transformer encoder), and DDPM (Diffusion probabilistic model), and provide a benchmark for future work in the area. 

\end{itemize}
\section{Related Work}\label{sec:related}
Our work, as it relates to prior research, draws from refractive lens distortion modeling, weather-induced artifact simulation, and deep learning-based image restoration. Below we organize prior related works in these three areas.

\subsection{Refractive Lens Distortions}\label{sec:refractive_lens}
The Brown-Conrady model \cite{brown1971close} is widely used in the synthetic generation of geometric lens distortions, and normally paired with Perlin noise \cite{perlin1985image}, Thin-Plate Spline methods \cite{bookstein1989principal}, or the Truncated-Power Law \cite{tatarskii1971turbulence} for more realistic refractive artifacts due to natural geometric irregularities and atmospheric turbulence at moderate computational cost. More recent lens distortion works, while more flexible and realistic through the use of deep learning to predict distortion parameters directly from images \cite{Feng2023SimFIR,liao2024DaFIR}, incur significantly higher computational costs, making them especially unsuitable for the low-resource settings we seek to address. Similarly, work by Li et al. \cite{li2024novel} proposed a divergence-free vector field formulation, which is effective but computationally expensive for atmospheric turbulence simulation in low-resource settings.

Our refractive lens distortion augmentation pipeline combines some of these methods, balancing computational efficiency and South African landscape scene realism in the context of resource-constrained autonomous perception scenarios.

\subsection{Weather-Induced Artifact Simulation}\label{sec:weather_induced_artifact_simulation}
Recent works in fog and haze simulation combine physics-inspired methods with deep learning for better realism and lower inference overhead. In particular, in \cite{li2023domain}, the authors integrate domain-adaptive object detection within synthetic fog environments, while \cite{zhang2024enhancing} unifies weather classification and object detection, demonstrating robust performance gains through computationally efficient augmentations. Other resource-intensive synthetic weather simulation approaches show that the use of Generative Adversarial Networks (GANs) or physics-based rendering engines can improve generalization in adverse weather conditions \cite{gupta2024robust}. Computational photography research continues to refine lens flare modeling. The Flare7K++ dataset \cite{wu2024flare7k} significantly advanced realistic flare generation, allowing networks to better generalize under adverse optical conditions. Such approaches incur moderate computational costs but deliver valuable realism crucial for autonomous driving scenarios, especially in challenging lighting conditions.

\subsection{Deep Learning-Based Restoration}\label{sec:deep_learning_image_restoration}
Early deep learning approaches for image restoration primarily used encoder–decoder architectures with skip connections, such as U-Net \cite{ronneberger2015u}.  Variants like U-Net++ and residual U-Nets further improved performance in low-resource settings \cite{zhou2019unetplusplus,Ramos2021ResnetUnet}. More recent Transformer-based architectures, such as Segformer \cite{xie2021segformer} and EvenFormer \cite{lu2025evenformer}, use lightweight or deformable attention mechanisms to improve efficiency and reconstruction quality.

Diffusion models, while computationally expensive to train, represent the state-of-the-art in image artifact correction and image generation \cite{ho2020denoising,Peebles2023DiT}, sometimes without the need for task-specific auxiliary loss functions \cite{saharia2022palette}. Some approaches, such as Diff-UNet \cite{xing2023diff}, improve efficiency by combining lightweight priors with diffusion, achieving competitive volumetric segmentation results in fewer steps. More recent work introduces parameter-efficient adapter modules \cite{hu2022lora} to large pre-trained diffusion models without retraining the full network \cite{liang2025diffusionadapter}, making scalable deployment of high-fidelity restoration models achievable in a low-resource setting without the high computation cost of training large diffusion models.

\section{Methodology}
\label{sec:method}
In this section, we describe our end-to-end data generation pipeline, which takes raw dashcam images from South African driving scenarios and generates a wide variety of refractive and weather-induced artifacts in the image. We first calibrate the camera using a standard ChArUco board to estimate intrinsic parameters and distortion coefficients, rectify the images with the calibrated parameters, then generate four different refractive and three weather-induced distortions based on the rectified images.
We provide three baseline models, a fine-tuned ImageNet-1k pre-trained ResNet34 U-Net, a SegFormer, and a diffusion-based image reconstruction model to remove the distortion and artifacts and evaluate the results. Figure \ref{fig:framework} shows our pipeline. We describe the data generation pipeline and the baseline restoration network in detail below.

\subsection{Refractive Distortions}\label{subsec:refractive}
Cheap dashcam lenses or atmospheric turbulence cause refractive optical distortion. By simulating these phenomena on undistorted frames, we can generate paired data (clean frame $\leftrightarrow$ distorted frame and the warping UV map) that closely resemble what an autonomous perception system would see on Southern African roads. We use OpenCV \cite{bradski2000opencv} and GSTools \cite{mueller2022gstools} to implement the UV warping vector field maps described below.

\captionsetup{skip=1pt}
\begin{figure*}[!thb]
  \centering
  \includegraphics[width=\textwidth,
  trim = 0 1.2cm 0 0,   
    clip]{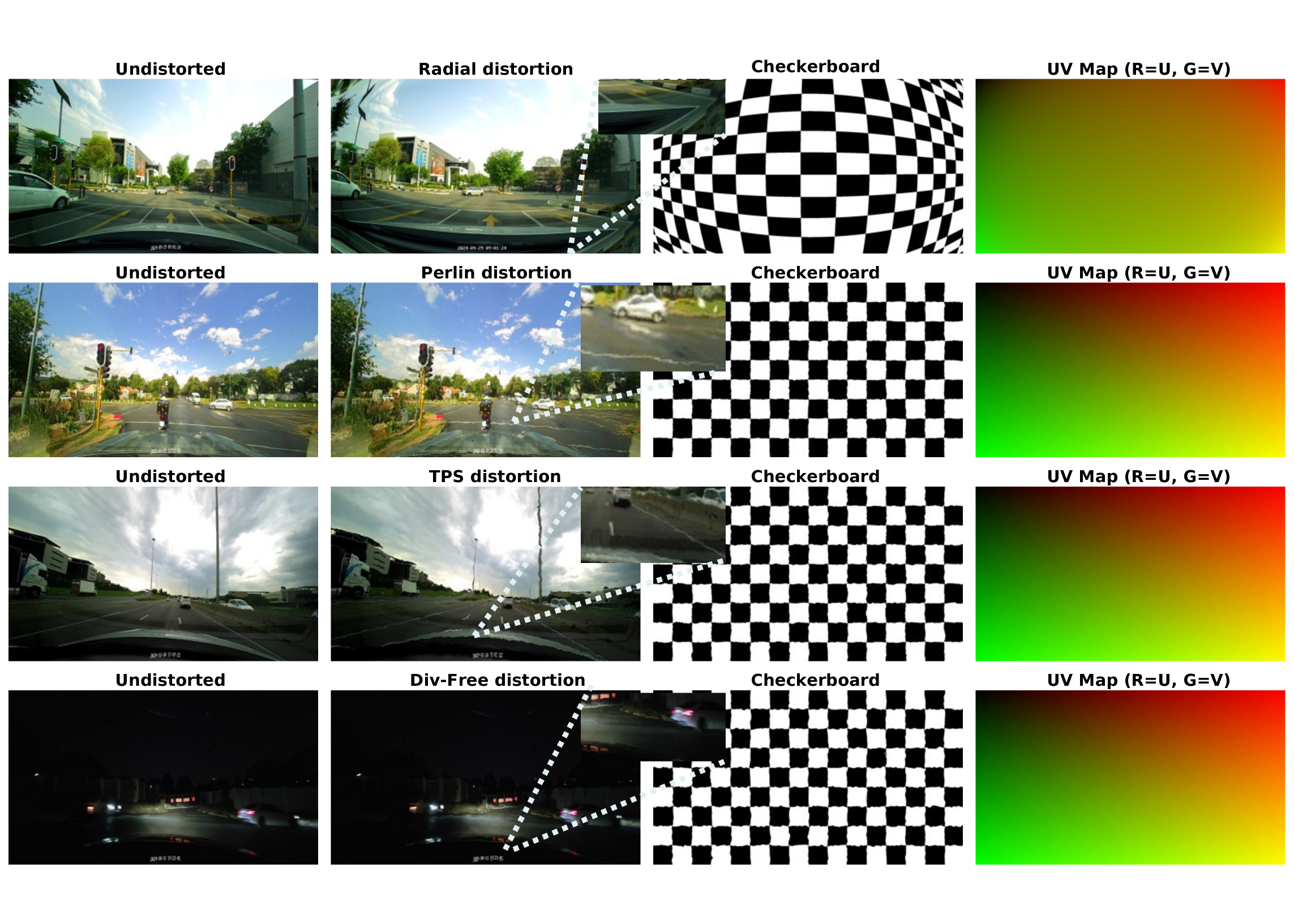}
  \caption{Visualization of refractive distortions for randomly selected samples: Columns (1) Undistorted frame, (2) Distorted frame, (3) Corresponding checkerboard pattern, and (4) Combined UV map with red channel representing $U$ and green channel representing $V$.}
  \label{fig:refractive_distortions}
\end{figure*}

\subsubsection{Radial Lens Distortion}
To simulate realistic dashcam lens distortions, we start with rectifying the original frame using the calibrated focal length and distortion, tangential and thin‐prism coefficients \(\{k_i,\,p_j,\,s_m\}\) from the Extended Brown–Conrady model. The rectified images will then be warped using random lens distortion parameters to mimic lens barrel and cushion distortion, misalignment, and manufacturing variability. In the normalized image plane \((x,y)\) (where \(x = X/Z\), \(y = Y/Z\), and \(r^2 = x^2 + y^2\)), we again use the Extended Brown–Conrady model below to distort the images, where the warped image coordinates becomes
\begin{align}
x_d &= x\left(1 + \sum_{i=1}^{6} k_i\,r^{2i} \right)
      + 2p_1 x y 
      + p_2(r^2 + 2x^2) \notag\\
    &\quad + s_1 r^2 + s_2 r^4,
\label{eq:brown_conrady_x} \\[4pt]
y_d &= y\left(1 + \sum_{i=1}^{6} k_i\,r^{2i} \right)
      + p_1(r^2 + 2y^2)
      + 2p_2 x y \notag\\
    &\quad + s_3 r^2 + s_4 r^4.
\label{eq:brown_conrady_y}
\end{align}
Applying this mapping, we distort the rectified image with UV offsets \(\bigl(\Delta x(x_d,y_d),\,\Delta y(x_d,y_d)\bigr)\) to mimic low-cost lens artifacts as depicted in Figure \ref{fig:refractive_distortions}, row 1.

\subsubsection{Perlin Distortion}
To simulate heat turbulence, we use Perlin noise-based \cite{perlin1985image} warping to mimic real‐world refractive turbulence or subtle surface irregularities. We sample two independent zero-mean Gaussian fields \(R_x, R_y\) with exponential covariance
\[
\mathrm{Cov}\bigl(R(\mathbf x),R(\mathbf x')\bigr)
= \exp\!\bigl(-\|\mathbf x-\mathbf x'\|/\ell\bigr),
\]
where \(\ell\) is the correlation length. After normalizing each field to unit variance \(\sigma_R\), we set pixel shifts
\begin{equation}
\Delta x(x,y) = \alpha\,\frac{R_x(x,y)}{\sigma_R}, 
\quad
\Delta y(x,y) = \alpha\,\frac{R_y(x,y)}{\sigma_R},
\end{equation}
with \(\alpha\), the maximum displacement (in pixels). Each source pixel \((x,y)\) is mapped to
\[
(x_d,y_d) = \bigl(x + \Delta x(x,y),\,y + \Delta y(x,y)\bigr),
\]
and we finally intepolate to form the final image.
Figure \ref{fig:refractive_distortions} row 2 shows an example of this distortion.

\subsubsection{Thin-Plate Spline (TPS) Distortion} 
To simulate bending from uneven surfaces or imperfect lens material in cheap dashboard-mounted cameras, we fit a Thin-Plate Spline (TPS) \cite{bookstein1989principal} to key-point pairs. Let \(\{(x_i,y_i)\}_{i=1}^N\) be source points and \(\{(u_i,v_i)\}_{i=1}^N\) their observed warped positions. We solve for  
\begin{equation}
f_x(x,y)
= a_{x,1} + a_{x,2}\,x + a_{x,3}\,y
  + \sum_{i=1}^N w_{x,i}\,U(r_i),
\end{equation}
\begin{equation}
r_i = \sqrt{(x - x_i)^2 + (y - y_i)^2},
\quad
U(r) = r^2 \ln r,
\end{equation}
and similarly for \(f_y\). The coefficients \(\{a_{x,1},a_{x,2},a_{x,3},w_{x,i}\}\) and \(\{a_{y,1},a_{y,2},a_{y,3},w_{y,i}\}\) are found by constraining  
$f_x(x_i,y_i)=u_i, f_y(x_i,y_i)=v_i$,
and \(
\sum_{i=1}^N w_{x,i}
=\sum_{i=1}^N w_{x,i}x_i
=\sum_{i=1}^N w_{x,i}y_i
=0,
\)
and similarly for \(w_{y,i}\). Each pixel \((x,y)\) is then mapped to  
\[
(x_d,y_d)=\bigl(f_x(x,y),\,f_y(x,y)\bigr),
\]
we then resample using
\(\bigl(x_d(x,y),\,y_d(x,y)\bigr)\) to get the distorting effect shown in Figure \ref{fig:refractive_distortions}, row 3.

\subsubsection{Incompressible (Divergence‐Free) Distortion} 
In addition, we use another warping map to diversify the refractive distortions further. We construct a divergence‐free vector field \cite{eisenberger2018divergence} via a scalar stream function \(\psi(x,y)\), ensuring $\nabla \cdot \mathbf{v}(x,y) \;=\; 0.$
Let
\begin{equation}
\label{eq:stream_function}
\mathbf{v}(x,y) \;=\; 
\begin{pmatrix}
u(x,y) \\ v(x,y)
\end{pmatrix}
\;=\; 
\begin{pmatrix}
\displaystyle \frac{\partial \psi(x,y)}{\partial y} \\[8pt]
\displaystyle -\,\frac{\partial \psi(x,y)}{\partial x}
\end{pmatrix}.
\end{equation}
By construction, 
\[
\frac{\partial u}{\partial x} \;+\; \frac{\partial v}{\partial y}
\;=\;
\frac{\partial}{\partial x}\Bigl(\tfrac{\partial \psi}{\partial y}\Bigr)
\;-\;
\frac{\partial}{\partial y}\Bigl(\tfrac{\partial \psi}{\partial x}\Bigr)
\;=\; 0.
\]
We sample \(\psi(x,y)\) as a smooth, zero‐mean random field (e.g.,\ a Gaussian random field with chosen correlation length), then compute \(u(x,y)\) and \(v(x,y)\) via \eqref{eq:stream_function}. Finally, each pixel \((x,y)\) in the undistorted frame is mapped to its warped location
\(
\label{eq:divfree_warp}
\bigl(x_d,\,y_d\bigr)
\;=\;
\Bigl(x \;+\; \alpha\,u(x,y),\quad y \;+\; \alpha\,v(x,y)\Bigr),
\)
where \(\alpha\) is a scaling factor (in pixels) controlling the maximum displacement. We then remap
\(\bigl(x_d(x,y),\,y_d(x,y)\bigr)\). The resulting image exhibits a smooth, swirl‐like deformation (zero divergence) that mimics wind‐driven or dust‐induced refractive warping. A checkerboard under this warp (by remapping each checker corner via \eqref{eq:divfree_warp}) visually confirms the divergence‐free flow pattern in Figure \ref{fig:refractive_distortions}, row 4.

\subsection{Weather-Induced Artifacts}\label{subsec:weather_induced}
For weather-induced artifacts, we additionally alter pixel intensities of the image, degrading visibility and image contrast without introducing geometric deformations. Below, we formally describe the three methods for weather-induced artifacts used in our pipeline.

\begin{figure*}[tbh]
  \centering
  \includegraphics[width=\textwidth]{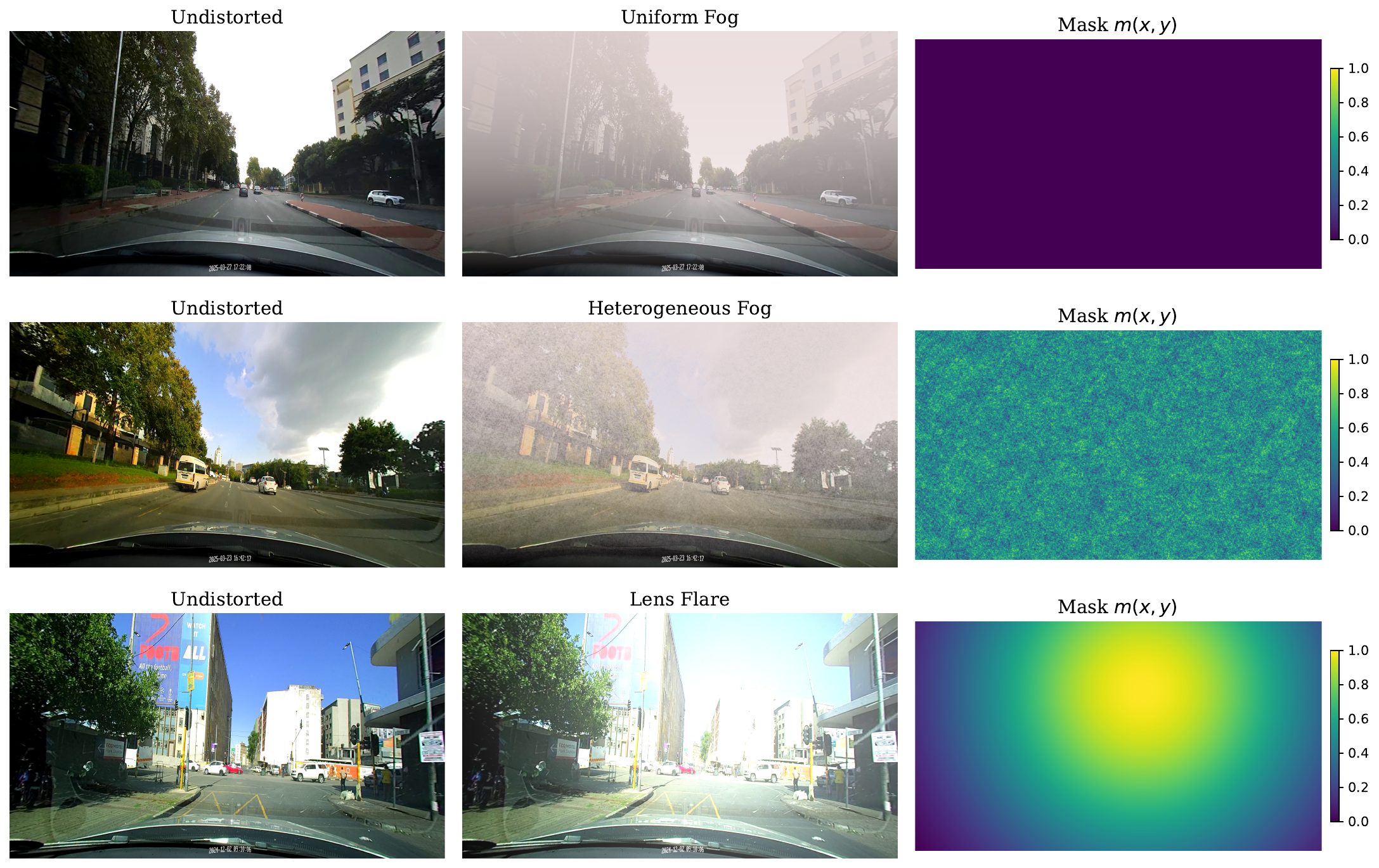}
  \caption{Example of the three simulated weather-induced artifacts (undistorted input, homogeneous fog, heterogeneous fog, and lens flare. Column 3 shows the mask of the same effect applied to a clear canvas.}
  \label{fig:heterofog}
\end{figure*}

\subsubsection{Uniform  Fog}\label{sec:uniform_fog}
Depth \(d(x,y)\) is approximated by a linear gradient from top to bottom of the image to simulate Uniform fog \cite{goovaerts1997geostatistics}:
\[
d(x,y) \;=\; \bigl(1 - \tfrac{y}{H}\bigr)\,D_{\max}, 
\quad
D_{\max} = 160.0 \text{ m,}
\]
where \(H\) is the image height. We choose a mean extinction coefficient 
\(
k_0 = \frac{-\ln(0.05)}{V_{\mathrm{vis}}} = \frac{-\ln(0.05)}{100\,\mathrm{m}} \approx 0.0375
\),
and perturb it by \(\pm5\%\) to obtain \(k = k_{0} \times (1 \pm 0.05)\). The fog‐transmission map is $t(x,y) \;=\; \exp\bigl(-\,k \, d(x,y)\bigr)\,$,and the final fogged pixel intensity $I_{\mathrm{fog}}(x,y)$ follows Koschmieder’s law:
\begin{align}
I_{\mathrm{fog}}(x,y) 
  &= I_{\mathrm{clean}}(x,y)\,t(x,y)
    + A\bigl(1 - t(x,y)\bigr)\,, 
  \label{eq:uniform_fog_line1} \\[2pt]
A 
  &= \begin{bmatrix}220 & 220 & 235\end{bmatrix}^\mathsf{T}\,,
  \label{eq:uniform_fog_line2}
\end{align}
where \(I_{\mathrm{clean}}(x,y)\) is the original (fog-free) RGB value and \(A\) is the atmospheric‐light RGB (in \([0,255]\)). We show an example of this artifact in Figure \ref{fig:heterofog}, row 1. 

\subsubsection{Heterogeneous Fog}\label{sec:heterogeneous_fog}
We vary the extinction coefficient \(k(x,y)\) through multi-depth‐scale Perlin noise as in \cite{zdrojewska2004fog},  with scales and values  shown in Table~\ref{tab:scales_weights}:

\begin{table}[h]
\centering
\begin{tabular}{ccccccc}
\hline
\(s_i\) & 4 & 8 & 16 & 32 & 64 & 128 \\ 
\(w_i\) & 0.30 & 0.22 & 0.15 & 0.11 & 0.08 & 0.07 \\ 
\hline
\end{tabular}
\caption{Depth-dependent noise scales and weights.}
\label{tab:scales_weights}
\end{table}

For each scale, find $P_i(x,y)$ and $P_{\mathrm{comb}}(x,y)$,
\begin{equation}
P_i(x,y)=\frac{\mathrm{Perlin}(x/s_i,y/s_i)-\min}{\max-\min}\in[0,1],
\label{eq:perlin_hetero_fog}
\end{equation}
\begin{equation}
P_{\mathrm{comb}}(x,y)=\frac{\sum_iw_i\,P_i(x,y)}{\sum_iw_i}.
\end{equation}
Enforce \(\mathbb{E}[k]=k_0\pm5\%\) by
\[
k(x,y)=k_0\,\frac{P_{\mathrm{comb}}(x,y)}{\langle P_{\mathrm{comb}}\rangle}\,(1+\delta),\quad|\delta|\le0.05,
\]
with \(d(x,y)\) as in Sec.~\ref{sec:uniform_fog}. Then

$t(x,y)=\exp\bigl(-k(x,y)\,d(x,y)\bigr)$
and
\begin{equation}
I_{\mathrm{het}}(x,y)=I_{\mathrm{clean}}(x,y)\,t(x,y)+A\bigl(1-t(x,y)\bigr).
\label{eg:het_flare_image}
\end{equation}
We then use \(k(x,y)\) as the heterogeneous \(k\)-map to apply per image as shown in Figure \ref{fig:heterofog}, row 2.

\subsubsection{Lens Flare Simulation}
To simulate dry-season sun glare in South African driving scenes \cite{cook2004wet}, we adopt a Gaussian flare model similar to \cite{lu2024glare}, placing the flare center \((c_x, c_y)\) randomly in the upper image region such that: 
\[
c_x \sim \mathcal{U}(0.3\,W,\,0.7\,W), 
\quad
c_y \sim \mathcal{U}(0,\,0.3\,H),
\]
where \(W\) and \(H\) are the image width and height.  Let the flare radius be
\[
r_{\mathrm{flare}} = \rho \,\sqrt{W^2 + H^2}, 
\quad
\rho \in [\,0.3 \pm 0.05\,]
\]
(perturbed by \(\pm5\%\)), and the intensity parameter $\beta \in [\,0.6 \pm 0.05\,]$. We use Euclidean pixel-to-flare center distance $d_f(x,y)$, and compute a normalized Gaussian mask:
\begin{equation}
m(x,y) = \exp\!\Bigl(-\tfrac12\bigl(\tfrac{d_f(x,y)}{r_{\mathrm{flare}}}\bigr)^2\Bigr)\in[0,1].
\end{equation}

The Flare $m(x,y) \in ([0,1])$ and image are combined, then clipped within valid intensity
\begin{equation}
I_{\mathrm{flare}}(x,y)
= \operatorname{clip}\bigl(I(x,y) + \beta\cdot255\,m(x,y),\,0,\,255\bigr),
\end{equation}

as shown in Figure \ref{fig:heterofog}, row 3.

\subsection{Dataset}\label{sec:Dataset}
The dataset consists of $1920 \times 1080$ RGB dashcam frames captured from casual driving in four southern African suburbs, namely Hatfield, Centurion, Midrand, and Sandton, during different times of the day and weather conditions that are not all equally represented in the data, as dashcams typically collect commutes from work every day as opposed to costly procedural data collection efforts. For this exercise, we randomly sample 16,000 images for refractive distortions and 20,000 for weather-induced artifacts from a much larger collection (over 300,000 images). For each selected frame, we sample transformations, controlled via configuration files. 
Our pipeline outputs pairs of undistorted and distorted images together with their corresponding warping UV maps (for refractive distortions) or intensity maps (for weather-induced artifacts).
In the following sections, we describe the experimental setup for training un-distortion and de-weathering models based on the synthetic data generated by the pipeline.


\section{Baseline Experiment Setup}\label{sec:setup}
We fine-tune three separate architectures from pre-trained weights for image reconstruction: ResUnet (ImageNet weights)\cite{Ramos2021ResnetUnet}, SegFormer-B1 (ImageNet weights)\cite{xie2021segformer}, and a Conditional DDPM Diffusion Model (ddpm-cat-256 weights) \cite{ho2020denoising}. We freeze the encoder weights and only allow decoder weight updates. Each model is trained independently for the two tasks: undistortion of refractive artifacts and deweathering of uniform fog, heterogeneous fog, and lens flare.

\subsection{Training Details}
We use an 80/20 random split for the training and validation sets, over 35 epochs for the refractive undistortion and 15 epochs for deweathering, with a batch size of 16. We use the AdamW optimizer with a learning rate of $5\times10^{-5}$ and ReduceLROnPlateau (factor 0.5, patience 5) under mixed‐precision AMP. For the conditional DDPM, we train on $256\times256$ crops for the same epoch counts with batch size 8. We use a linear beta DDPMS scheduler ($T=1000$), AdamW (LR $1\times10^{-5}$) with a cosine warm-up schedule (500 warm-up steps) and mixed-precision AMP. 

\textbf{Refractive Undistortion} supervision uses ground‐truth UV maps and clean RGB targets such that the  ResUnet and SegFormer loss function is given by:

\[
\mathcal{L} = \| \hat{I} - \mathrm{GT}_{\mathrm{rgb}} \|_1 + \| \hat{y}_{\mathrm{uv}} - \mathrm{GT}_{\mathrm{uv}} \|_1,
\]
where \( \hat{y}_{\mathrm{uv}} \) is the predicted UV flow, \( \mathrm{GT}_{\mathrm{uv}} \) is the ground truth UV flow, and \( \mathrm{GT}_{\mathrm{rgb}} \) is the clean RGB target. The predicted image \( \hat{I} \) is recovered by warping the distorted input with the predicted flow:
$\hat{I} = \mathrm{warp}(x,\, \hat{y}_{\mathrm{uv}}).$
where \( x \) is the distorted input image. The DDPM Loss is given by:
\begin{equation}
    \mathcal{L}_{\mathrm{DDPM}}
    = 0.8\,\|\hat\epsilon - \epsilon\|_{1}
    + \|\mathrm{warp}(x,\,\hat y) - \mathrm{GT}_{\mathrm{rgb}}\|_{2}^{2}.
\end{equation}
The first term trains a UV-denoising U-Net, and the second term pushes the warped output towards the ground truth image. The \textbf{Deweathering} objective is given by:
\[
\mathcal{L} = \| \hat{I} - \mathrm{GT}_{\mathrm{rgb}} \|_1
\]
where \( x \) is the corrupted input image, \( \hat{I} \) is the predicted deweathered output, and \( \mathrm{GT}_{\mathrm{rgb}} \) is the ground truth clean image.

\subsection{Evaluation Metrics}\label{sec:evaluation}

\begin{figure*}[thb]
  \centering
  \includegraphics[width=\textwidth,
                    height=0.25\textheight,%
                    keepaspectratio
  ]{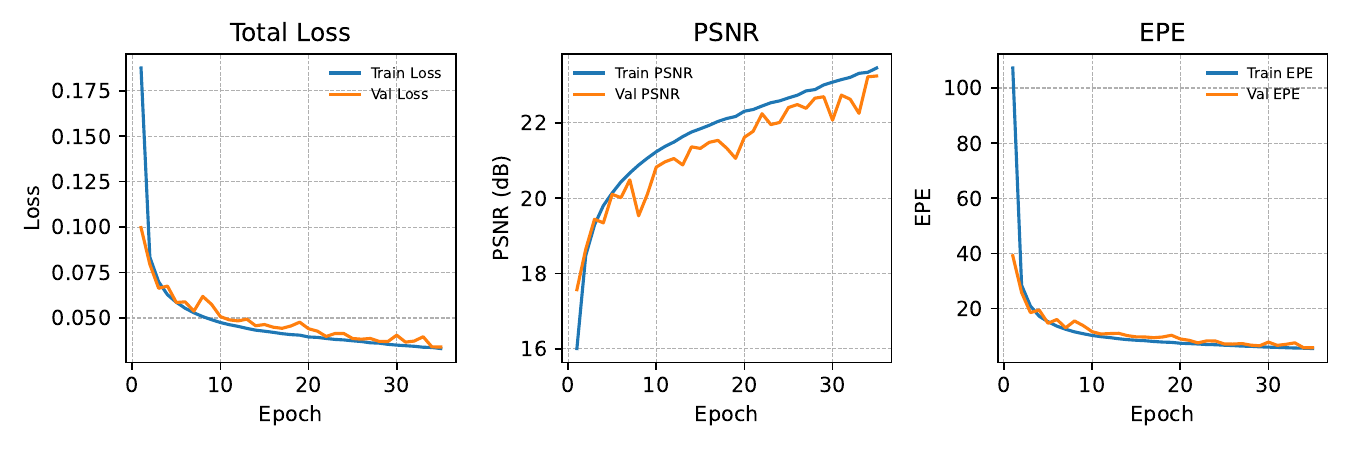}
  \caption{
    \textbf{Refractive Distortion Training and Validation Curves.}
    (\textbf{a}) Total Loss (train vs.\ val), showing steady convergence.
    (\textbf{b}) PSNR improvement over epochs for both train and val.
    (\textbf{c}) Endpoint Error (EPE) on UV flow decreases as training proceeds.
  }
  \label{fig:geom_metrics}
\end{figure*}

\begin{figure*}[thb]
  \centering
  \includegraphics[width=\textwidth,
                    height=0.25\textheight,%
                    keepaspectratio]{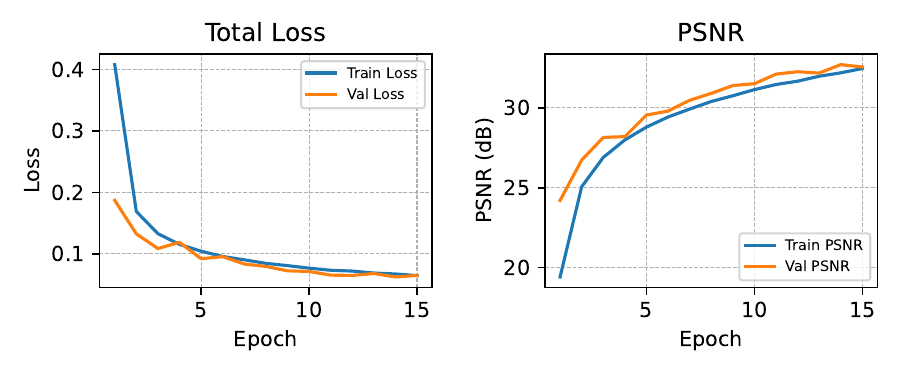}
  \caption{
    \textbf{Synthetic Weather-induced Artifact Training and Validation Curves.}
    (\textbf{a}) Total loss (train vs.\ val).
    (\textbf{b}) PSNR improvement over epochs for both train and val.
  }
  \label{fig:photo_metrics}
\end{figure*}

We report both Peak Signal-to-Noise Ratio (PSNR) for refractive undistortion and de-weathering result images and Endpoint Error (EPE) for refractive undistortion UV map:
\[
\text{PSNR} = 10 \cdot \log_{10} \left( \frac{\text{MAX}^2}{\text{MSE}} \right), \text{EPE} = \frac{1}{N} \sum_{i=1}^{N} \left\| \hat{y}_{\mathrm{uv}}^{(i)} - \mathrm{GT}_{\mathrm{uv}}^{(i)} \right\|_2
\]
For PSNR, MAX = 255, and the mean square error (MSE) is computed between the predicted and ground truth images. For UV map EPE, \( N \) is the number of pixels and \( \hat{y}_{\mathrm{uv}}^{(i)} \) and \( \mathrm{GT}_{\mathrm{uv}}^{(i)} \) are the predicted and true flow vectors at pixel \( i \), respectively.

\section{Results and Discussion}\label{sec:results}

As shown in Figures~\ref{fig:geom_metrics} and \ref{fig:photo_metrics}, the refraction undistortion model actually minimizes its combined loss more quickly: both train and validation losses fall below 0.075 within the first five epochs, whereas the weather‐correction model only reaches that level around epoch 10.  However, the PSNR trajectories tell a different story: by epoch 5, the Deweathering model has already climbed above 28 dB (Fig.~\ref{fig:photo_metrics}b), while the refractive model only reaches 20 dB at the same point.  This suggests that, although the refractive undistortion model finds a good minimum of its combined loss faster, the photometric task yields larger perceptual gains per epoch.

\begin{figure*}[thb]
  \centering
  \includegraphics[width=\textwidth,trim = 0 3cm 0 0,   
    clip]{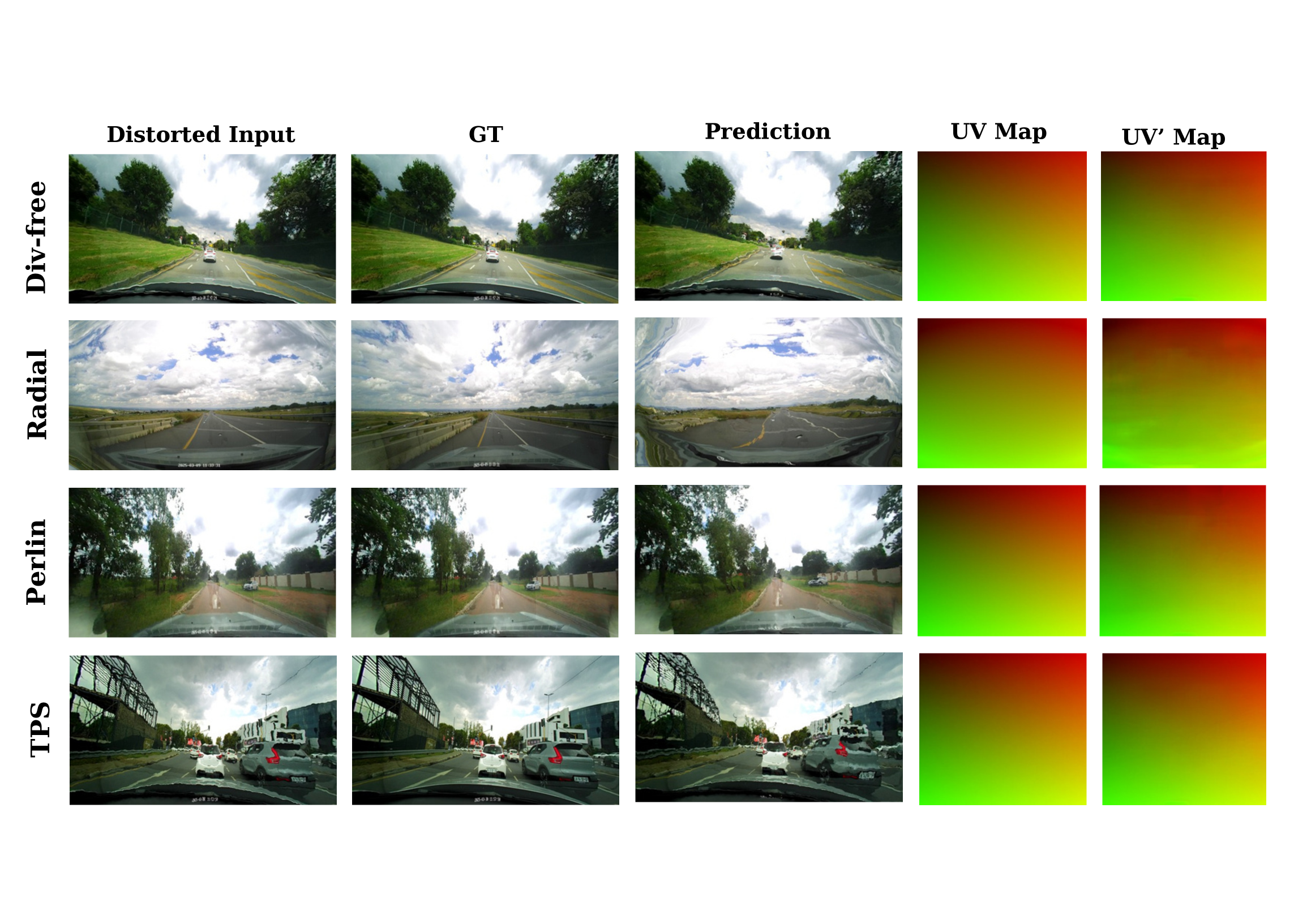}
  \caption{ Comparison of the undistorted input image, distorted image, and the best model's undistortion of the input for all four types of refractive distortions we cover.}
  \label{fig:output_refractive_undistortions}
\end{figure*}

\begin{figure*}[thb]
  \centering
  \includegraphics[width=0.9\textwidth]{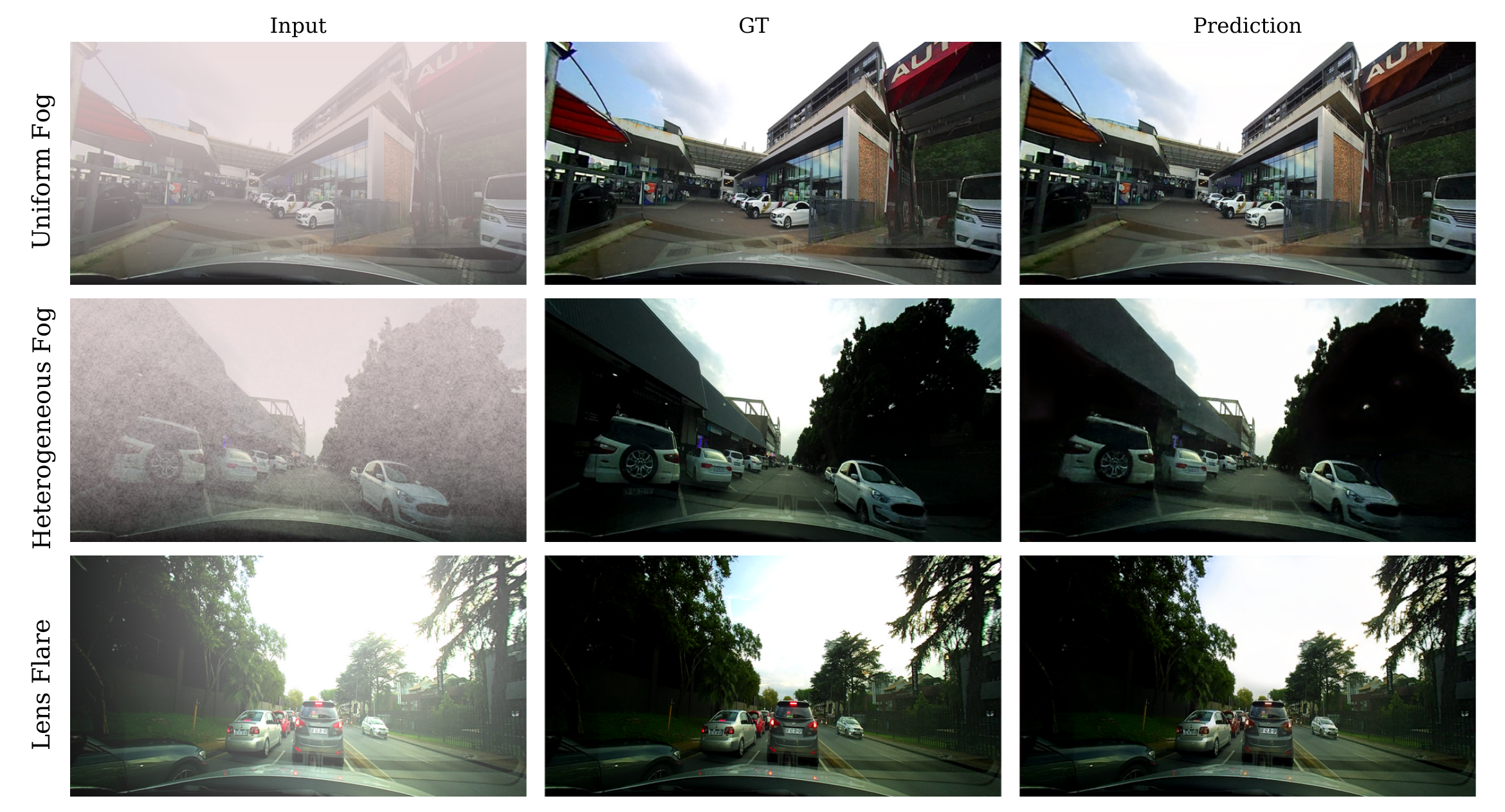}
  \caption{ Qualitative deweathering results for three weather-induced artifacts: uniform fog, heterogeneous fog, and lens flare. From left to right: degraded input, ground truth, and prediction.}
  \label{fig:output_deweathering_artifacts}
\end{figure*}

Table~\ref{tab:final_metrics} reports quantitative validation metrics for both weather artifacts ($W$) and refractive distortion ($R$) restoration tasks.  On the weather-induced artifacts, the ResUnet backbone (40M parameters) achieves the highest PSNR (32.54 dB) despite having a slightly larger combined loss $\mathcal{L}$ than the diffusion model (35M parameters). We find that the lighter SegFormer-b1 (14M parameters) struggles comparatively and likely underfits. It is possible that the larger SegFormer-b1 variants (b2-b5) would post much better performance.
On the refractive distortion side, both ResUnet and DDPM perform comparatively the same in terms of PSNR and EPE, while the smaller SegFormer performs the poorest, as shown in Table~\ref{tab:final_metrics}. As would be expected, the gap between the best and worst models seems to widen on the harder refractive undistortion task. In both tasks, a lower loss of the UV map $\mathcal{L}_{k}$ is strongly correlated with a reduced EPE, confirming that accurate UV map prediction is critical for distortion, especially in the case of strong distortions of the radial lens.  Figures \ref{fig:output_refractive_undistortions} and \ref{fig:output_deweathering_artifacts} show that deweathering yields almost perfect reconstructions, whereas refractive undistortion still leaves visible residual warps, especially for radial lens distortion.

\captionsetup{skip=2pt}

\setlength{\textfloatsep}{5pt}       
\setlength{\floatsep}{6pt}           
\setlength{\abovecaptionskip}{3pt}   
\setlength{\belowcaptionskip}{3pt}   
\begin{table}[tbh]
\centering
\begin{tabular}{lccccc}
\toprule
\textbf{Model}  & $\boldsymbol{\mathcal{L}}$ & $\boldsymbol{\mathcal{L}_{\mathrm{rec}}}$ & $\boldsymbol{\mathcal{L}_{k}}$ & \textbf{PSNR} & \textbf{EPE} \\
\midrule\midrule
$W$ ResUnet     & 0.065                      & 0.059                                   & 0.007                         & 32.54        & -            \\
$W$ SegFormer     &  0.096                     & 0.086                                   & 0.010                         & 28.99        & -           \\
$W$ DDPM     &  0.054                     & 0.051                                   & 0.0065 & 30.80        & -            \\
\hline
$R$ ResUnet       & 0.034                     & 0.029                                  & 0.005                         & 23.24        & 5.76         \\
$R$ SegFormer       & 0.146                     & 0.085                                  & 0.055                         & 16.29       & 69.82         \\
$R$ DDPM       &  0.044                    & 0.039                                 & 0.0048                         & 22.80        &   6.01       \\
\bottomrule
\end{tabular}
\caption{Validation metrics for the refractive $R$ and weather-induced artifact $W$ image restoration. Here, $\mathcal{L}$ is the combined weighted loss, $\mathcal{L}_{\mathrm{rec}}$ the L1 reconstruction loss, $\mathcal{L}_{k}$ the L1 k‐map/UV loss, PSNR, and EPE as described in Section \ref{sec:evaluation}.}
\label{tab:final_metrics}
\end{table}

\section{Conclusion}
\label{sec:conclusion}
This work takes a crucial step toward democratizing autonomous driving research by addressing the data scarcity and environmental variability prevalent in underrepresented regions such as Africa. By introducing a lightweight, scalable pipeline for simulating refractive distortions and weather-induced artifacts on real dashcam footage, we enable dataset expansion without reliance on expensive hardware or simulation infrastructure. Our baseline methods and recovery benchmarks establish a foundation for future research in image restoration and robust perception under challenging conditions. Ultimately, this effort aims to make autonomous driving technologies more inclusive and adaptable to diverse global contexts.

\section{Acknowledgments}
\label{sec:acks}
We acknowledge technical support and advanced computing resources from the University of Hawaii Information Technology Services – Research Cyberinfrastructure, funded in part by National Science Foundation CC* awards \texttt{\#}2201428 and \texttt{\#}2232862.

{
    \small
    \bibliographystyle{ieeenat_fullname}
    \bibliography{main}
}

\end{document}